\documentclass{article}

\usepackage{microtype}
\usepackage{graphicx}
\usepackage{subfigure}
\usepackage{booktabs} 
\usepackage{amsmath}
\usepackage{breqn}
\usepackage{listings}
\usepackage[title]{appendix}
\usepackage{xcolor}  	

\usepackage{hyperref}

\usepackage[preprint]{nips}
\definecolor{commentcolor}{RGB}{85,139,78}
\definecolor{stringcolor}{RGB}{206,145,108}
\definecolor{keywordcolor}{RGB}{34,34,250}
\definecolor{backcolor}{RGB}{245,245,244}

\usepackage{accsupp}	
\usepackage{enumitem}

\usepackage{listings}
\title{DeepRobust: A PyTorch Library for Adversarial Attacks and Defenses}

\author{
  Yaxin Li\\
  Michigan State University\\
  \texttt{liyaxin1@msu.edu}\\
  \And
  Wei Jin\\
  Michigan State University\\
  \texttt{jinwei2@msu.edu}\\
  \AND
  Han Xu\\
  Michigan State University\\
  \texttt{xuhan1@msu.edu}\\
  \And
  Jiliang Tang\\
  Michigan State University\\
  \texttt{tangjili@msu.edu}
}

\begin{document}

\maketitle

\begin{abstract}
DeepRobust is a PyTorch~\cite{paszke2019pytorch} adversarial learning library which aims to build a comprehensive and easy-to-use platform to foster this research field. It currently contains more than 10 attack algorithms and 8 defense algorithms in image domain and 9 attack algorithms and 4 defense algorithms in graph domain, under a variety of deep learning architectures. In this manual, we introduce the main contents of DeepRobust with detailed instructions. The library is kept updated and can be found at
\textit{\url{https://github.com/DSE-MSU/DeepRobust}}.
\end{abstract}

\section{Introduction}

\begin{figure}[htbp]

\centering
\subfigure[Image Adversarial Example]{
\begin{minipage}[t]{0.45\linewidth}
\centering
\includegraphics[width=2.5in]{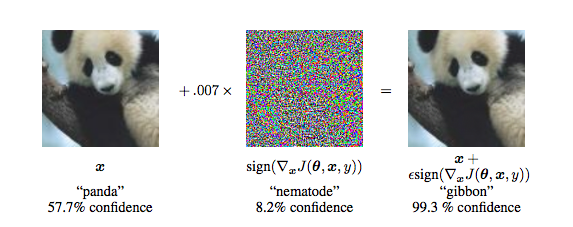}
\label{img_ex}
\end{minipage}%
}%
\subfigure[Graph Adversarial Example]{
\begin{minipage}[t]{0.45\linewidth}
\centering
\includegraphics[width=2.5in]{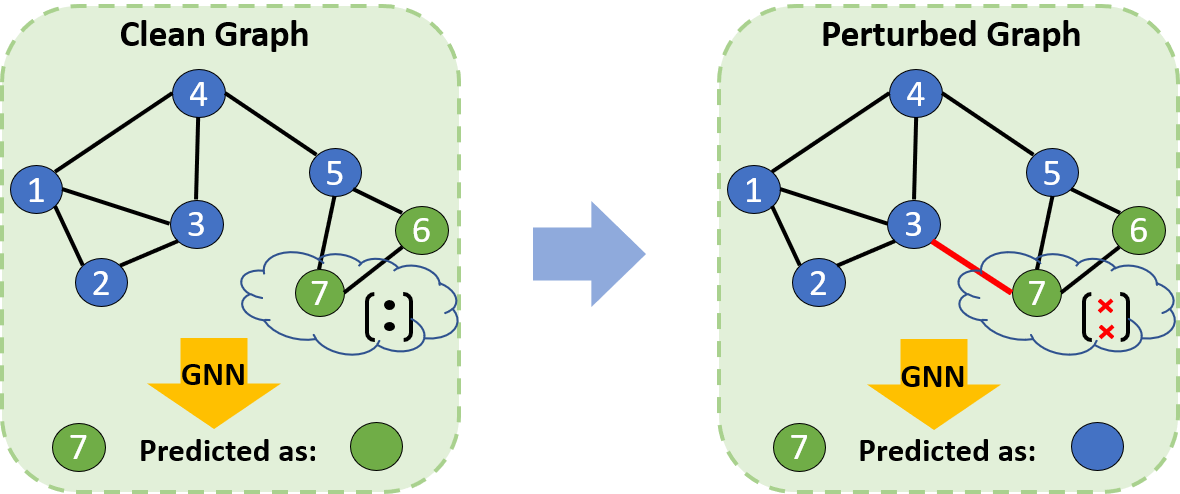}
\end{minipage}%
\label{gra_ex}
}%
\centering
\caption{An Illustration of Adversarial Example.}
\end{figure}

Deep learning has advanced numerous machine learning tasks such as image classification, speech recognition and graph representation learning. Since deep learning has been increasingly adopted by real-world safety-critical applications such as autonomous driving, healthcare and education, it is crucial to examine its vulnerability and safety issues. Szegedy et al.  \cite{szegedy2013intriguing} first found that Deep Neural Networks (DNNs) are vulnerable to small designed perturbations, which are called adversarial perturbations. Figure \ref{img_ex} demonstrated adversarial example in the domains of images and graphs. Since then, tremendous efforts have been made on developing attack methods to fool DNNs and designing their counter measures. As a result, there is a growing need to build a comprehensive platform for adversarial attacks and defenses. Such platform enables us to systematically experiment on existing algorithms and efficiently test new algorithms that can deepen our understandings, and thus immensely foster this research field.

Currently there are some existing libraries in this research field, such as \textit{Cleverhans}~\cite{papernot2018cleverhans}, \textit{advertorch}~\cite{ding2019advertorch}. They mainly focused on attack methods in the image domain. However, little attention has been paid on defense methods. Furthermore, the majority of them are dedicated to the image domain while largely ignoring other domains such as graph-structured data. Our library \textit{DeepRobust} not only provides representative attack and defense methods in the image domain, but also covers algorithms for graph data. This repository contains most classic and advanced algorithms

The remaining of this report is organized as follows. Section 2 introduces key concepts for adversarial attacks and defenses. Section 3 gives an overview of the DeepRobust library. Sections 4 and 5 introduce math background and implementation details of the algorithms in the library for the image and graph domains, separately. Section 6 provides concrete examples to demonstrate how to use the library. 

\section{Foundations of Adversarial Attacks and Defenses}
The main goal of attack algorithms is to make invisible perturbations on data and then lead to wrong classification of a classifier which normally has good performance on clean data. The study of attacks can be categorized from different perspectives such as attackers' goal, attackers' ability and so on. 

According to attackers' goal, attack methods can be categorized as follows:
\begin{itemize}[leftmargin = 20pt]
    \item {\bf Poisoning Attack} vs. {\bf Evasion Attack}\\
    {\bf Poisoning Attack:} Poisoning attacks refer to the attacking algorithms that allow an attacker to insert/modify several fake samples into the training data of a DNN algorithm. This training process with fake samples can cause bad performance on the test data. 
    
    {\bf Evasion Attack:} For evasion attack, the victim classifiers are fixed and normally have good performance on benign test samples. The adversaries do not have authority to change the classifier or its parameters, instead they craft some fake samples that the classifier can neither correctly classify nor distinguish them as an unusual input. In other words, the adversaries generate some fraudulent examples to evade detection by the classifier. 
    
    \item {\bf Targeted Attack} vs. {\bf Non-Targeted Attack}\\
    {\bf Targeted attack:} When the victim sample $(x,y)$ is given,  where $x$ is  feature  vector  and $y\in Y$ is  the ground truth label of $x$, the adversary aims to induce the classifier to give a specific label $t\in Y$ to the perturbed sample $x'$. 
    
    {\bf Non-Targeted Attack:} If there is no specific $t$ given, an adversarial example can be viewed a successful attack as long as it is classified as any wrong label.
\end{itemize}    

According to attackers' ability, attack methods can be grouped as follows:
\begin{itemize}[leftmargin = 20pt]
   \item {\bf White-Box attack:} the adversary has access to all the information of the target model, including its architecture, parameters, gradients, etc. 
    
   \item {\bf Black-Box attack:} In a black-box attack setting, the inner configuration of the target model is unavailable to adversaries. This type of methods often based on a bunch of attack queries.
    
   \item {\bf Grey-Box attack:} In a gray box attack setting, the attacker trains a generative model for producing adversarial examples in a white-box setting. Once the generative model is trained, it can be used to craft adversarial examples in a black-box setting.
\end{itemize}  

To mitigate the risk of adversarial attack, different countermeasures have been investigated. There are four main categories of defenses. The first one is robustness optimization, that is to do adversarial training, namely to use adversarial examples to retrain the model. Another type is adversarial example detection. The goal is to distinguish adversarial examples from data distribution. The third type is gradient masking. This type of defense  mainly includes some pre-processing methods to hide the gradient, in order to make the optimization in attack process much harder. The last type, provable defense, has gradually becomes a important stream of defense. Those methods aims to provide adversarial robustness guarantee.

\section{An Overview of DeepRobust} 


In this section, we aim to first provide an overview of the DeepRobust library including environment requirements and the overview design of DeepRobust. 

\subsection{Environment Requirements and Setup}
DeepRobust works on python $\geq 3.6$ and pytorch $\geq 1.2.0$. All dependencies are listed in Appendix A. After downloading this repository, run {\bf setup.py} to install DeepRobust into local python environment.

\subsection{The Overview Design}
This repository mainly includes two components -- the image component and the graph component as below. The directory structure can be found in Appendix B.
\begin{itemize}
    \item Image package
        \begin{itemize}
            \item attack
            \item defense
            \item netmodels
            \item evaluation
            \item configs
        \end{itemize}
    \item Graph package
        \begin{itemize}
            \item targeted\_attack
            \item global\_attack
            \item defense
            \item data
    \end{itemize}
\end{itemize}

{\bf The Image Component:}
According to the function of each program, the image component is divided into several sub-packages and contents. \textit {Attack} sub-package includes attack base class and attack algorithms. \textit {Defense} sub-package contains defense base class and defense algorithms. In {\bf Section 3}, we will give specific introduction for each algorithm. \textit {Netmodels} contains different network model classes. Users can simply generate a victim model by instantiating one model class. Through the evaluation program, we provide an easy-to-use API to test attack against defense. All the default parameters are saved in \textit {configs}.

{\bf The Graph Component: }The graph component contains several sub-packages and contents based on the functions. \textit{Targeted-attack} sub-package includes the targeted attack base class and famous targeted attack algorithms. Similarly, global attack base class and global attack algorithms are included in \textit{global-attack} sub-package. \textit{Defense} sub-package contains GCN model and other methods for defending graph adversarial attack. Besides, sub-package \textit{data} provides an easy access to public benchmark datasets including Cora, Cora-ml, Citeseer, Polblogs and Pubmed as well as pre-attacked graph data.

\section{Image Component}
In this section, we aim to give an introduction of the interface for attack and defense methods in the image domain. Meanwhile, we provide algorithm description and implementation details for each algorithm. A comprehensive survey about attack and defense methods in the image domain can be found in~\cite{xu2019adversarial}. 

\subsection{Attacks}
This subsection introduces the API for the attack methods in the image domain. Currently, this package covers nine representative attack algorithms including LBFGS\cite{szegedy2013intriguing}, FGSM\cite{goodfellow2014explaining}, PGD\cite{madry2017towards}, CW\cite{carlini2017towards}, onepixel\cite{su2019one}, deepfool\cite{moosavi2016deepfool}, BPDA\cite{athalye2018obfuscated}, Universal\cite{moosavi2017universal} and Nattack\cite{li2019nattack}.

For these algorithms, we currently support the following neural networks and datasets:
\vskip 0.1in
Supported networks:
\begin{itemize}
    \item CNN
    \item ResNet-18/34
    \item DenseNet
    \item VGG-11/13/16/19
\end{itemize}

Supported datasets:
\begin{itemize}
    \item MNIST
    \item CIFAR10
\end{itemize}

\subsubsection{Attack Base Class}

\textbf{deeprobust.image.attack.base\_attack}

In order to make further development flexible and extendable, we organize functions shared by different methods in one module as the attack base class. The main body of the algorithm is override in each subclass. Following are detailed instructions for the functions contained in this class:

\begin{itemize}
    \item \textbf{\_\_init\_\_}(self, model, device = 'cuda') 
    
    Initialization is completed in this function.
    \\[6pt]
    Parameters:
    \begin{itemize}
        \item model: the victim model.
        \item device: whether the program is run on GPU or CPU.
    \end{itemize}
    
    \item \textbf{check\_type\_device}(self, image, label, **kwargs) 
    
    The main purpose for this function is to convert the input into a unified data type so that they can be correctly used in the algorithm procedure.
    \\[6pt]
    Parameters:
    \begin{itemize}
        \item image: clean input.
        \item label: ground truth label corresponding to the clean input.
        \item **kwargs: optional input dependent on each derived class.
    \end{itemize}
    
    \item \textbf{parse\_params}(self, **kwargs)
    
    This function provides the interface for these user defined parameters.
    \\[6pt]
    Parameters:
    \begin{itemize}
        \item **kwargs: optional input dependent on each derived class.
    \end{itemize}
    \item \textbf{generate}(self, image, label, **kwargs)
    
    Call \textbf{generate()} to launch the attack algorithms.
    \\[6pt]
    Parameters:
    \begin{itemize}
        \item **kwargs: optional input. Parameters for the attack algorithms.
    \end{itemize}

\end{itemize}

\subsubsection{Attack algorithms}
\textbf{deeprobust.image.attack.lbfgs} \ L-BFGS attack\cite{szegedy2013intriguing} is the key work that has arisen people's attention on the neural network's vulnerability to small perturbation. This work tries to find a minimal distorted adversarial example by solving a intuitive box-constraint optimization problem: 
\begin{equation}
    \begin{split}
        &\text{minimize } ||x'-x||_2^2 \\
        &\text{subject to } C(x') = t \text{ and } x'\in[0,1]^m
    \end{split}
\end{equation}
where $x'$ is the adversarial example and $x$ is the clean example. We aim to find an adversarial example which could be classified as certain class and is close to the clean image. From the implementation perspective, addressing this optimization problem with two constraints is hard. Thus, this work turns to solve an alternative problem by doing binary search on a parameter $c$ to balance the trade off between perturb constraint and attack success for target class, and then use the LBFGS algorithm to get an approximate solution: 
 \begin{equation}
    \begin{split}
        &\text{minimize } c||x'-x||_2^2 + L(\theta, x', t) \\
        &\text{subject to } x'\in[0,1]^m
    \end{split}
\end{equation}
where $L(\theta, x', t)$ denotes the loss value of $x'$ for the target class t.

~\\
\textbf{deeprobust.image.attack.fgsm} \ Fast Gradient Sign Method(FGSM)\cite{goodfellow2014explaining} is an one-step optimization problem. The intuition is to move the input sample along the gradient direction to achieve highest loss value corresponding to the ground truth label. Thus the perturbed samples could fool the network with high confidence. To guarantee that the adversarial example lies in a small nearby area of starting the point $x$, the gradient descent step is followed by a \textit{clip} operation. Thus, the formulation of the process is described as follows:
\begin{equation}
    x' = Clip_{x,\epsilon}(x + \epsilon sign(\nabla_x L(\theta, x, y)))
\end{equation}
Here, \textit{Clip} denotes a function to project its argument to the surface of $x$'s $\epsilon$-neighbor ball. 

~\\
\textbf{deeprobust.image.attack.pgd} \ Projected Gradient Descent(PGD)\cite{madry2017towards} is an iterative version of the FGSM attack. The formulation of generating $x'$ is: 
\begin{equation}
    \begin{split}
        &x_0 = x \\
        &x_{t+1} = Clip_{x,\epsilon}(x^t + \alpha sign(\nabla_x L(\theta, x^t, y)))
    \end{split}
\end{equation}
It chooses the origin image as a starting point. PGD attack could create strong adversarial examples and is often used as a baseline attack for defense methods.

~\\
\textbf{deeprobust.image.attack.deepfool} \ Deepfool attack aims to find a shortest path to let the data point $x$ go across the decision boundary. It starts from an binary classifier. Denote the hyperplane as $F = \{x : w^T + b = 0\}$. The minimum perturbation is the distance from the data point $x_0$ to the hyperplane that can be: 
\begin{equation}
r^*(x_0) = -\frac{f(x)}{||w||^2}w    
\end{equation}
This calculation can be extended to general classifiers and also extend this problem to $l_p$ norm constraint perturbation. Compare to other methods like FGSM and PGD, Deepfool attack produces less perturbation to attack successfully. 

~\\
\textbf{deeprobust.image.attack.cw} \ Carlini and Wagner's attack ~\cite{carlini2017towards} aims to solve the same problem as defined in L-BFGS attack, namely trying to find the minimally-distorted perturbation. 

It addresses the problem by instead solving:
\begin{equation}
        \text{minimize  } ||x-x'||_2^2+c\cdot f(x',t)\\
        \text{ subject to  } x' \in [0,1]^m
\end{equation}
where $f$ is defined as $f(x',t) = (\max_{i\neq t}{Z(x')_i}-Z(x')_t)^+$. Here, $Z()$ is the logits function. Minimizing $f(x',t)$ encourages the algorithm to find an $x'$ that has the larger score for class $t$ than any other label, so that the classifier will predict $x'$ as the class $t$. Next, by applying a line search on the constant $c$, it can find the $x'$ that has the least distance to $x$. 

~\\
\textbf{deeprobust.image.attack.universal} \ Previous methods only consider one specific targeted victim sample $x$. However, the work ~\cite{moosavi2017universal} devises an algorithm that successfully misleads a classifier's decision on almost all test images. It tries to find a perturbation $\delta$ satisfying:
\begin{enumerate}
  \item $||\delta||_p \leq \epsilon$.\\
  \item $\underset{x \sim D(x)}{{P}}   (f(x+\delta) \neq f(x))\geq 1-\sigma$.
\end{enumerate}
Formulation 1 is constraint of the norm of perturbation size and formulation 2 set a threshold for the probability of misclassification. Actually it aims to find a invisible perturbation $\delta$ such that the classifier gives wrong decisions on most of the samples. 

~\\
\textbf{deeprobust.image.attack.onepixel} \ One pixel attack ~\cite{Su_2019} constraints the perturbation by the $l_0$ norm instead of $l_2$ norm or $l_\infty$ norm. That is to say, finding the minimum number of pixels to perturb.
Differential evolutionary algorithm(DE) is applied to solve this optimization problem.

~\\
\textbf{deeprobust.image.attack.bpda} \ Backward Pass Differentiable
Approximation(BPDA)~\cite{athalye2018obfuscated} is a technique to attack defenses where gradients are not readily available. BPDA solves this problem by finding an approximation function for the non-differential layer and calculating the gradient through the approximation function.

~\\
\textbf{deeprobust.image.attack.nattack} \ Nattack ~\cite{li2019nattack} is a black box attack  trying to find a probability density distribution of where adversarial samples lying over a small region centered around the input. One sample that are drawn from this distribution is likely to be an adversarial example. 

It uses an intuitive way to find this density distribution. First, it initializes the distribution with random parameters. Then, it samples from this distribution several times as the neural network input and calculates the average loss value of those samples. Thus, the average loss is a function of the distribution parameters. Finally, It performs gradient decent on the average loss and updates the distribution parameters. It will iterates this process until successful attack.

\subsection{Defense Subpackage}
This subsection introduces the API for the defense methods in the image domain. Until now, this package covers three categories of defense methods including adversarial training, gradient masking and detection.

\subsubsection{Defense Base Class}
\textbf{deeprobust.image.defense.base\_defense}

This module is the base class of all the adversarial training algorithms. It provides basic components for defense methods. Following functions are contained in this class:
\begin{itemize}
    \item \textbf{\_\_init\_\_}(self, model, device)
    
    Parameter initialization is completed in this function.
    \\[6pt]
    Parameters:
    \begin{itemize}
        \item model: the attack victim model.
        \item device: whether the program is run on GPU or CPU.
    \end{itemize}
    
    \item \textbf{parse\_params}(self, **kwargs)
    
    This function provides the interface for user defined parameters.
    \\[6pt]
    Parameters:
    \begin{itemize}
        \item **kwargs: optional input dependent on each derived class.
    \end{itemize}
    
    \item \textbf{generate}(self, train\_loader, test\_loader, **kwargs)
    
    Call \textbf{generate()} to launch the defense.
    \\[6pt]
    Parameters:
    \begin{itemize}
        \item **kwargs: optional input dependent on each derived class.
    \end{itemize}
    
    \item \textbf{loss}(self, output, target)
    
    Calculate the training loss. This function will be overridden by each defense class according to the algorithms requirements.
    \\[6pt]
    Parameters:
    \begin{itemize}
        \item output: model output.
        \item target: ground truth label.
    \end{itemize}
    
    \item \textbf{adv\_data}(self, model, data, target, **kwargs)
    
    Generate adversarial training samples for robust training. This function will be overridden by each defense class according to algorithms requirements.
    \\[6pt]
    Parameters:
    \begin{itemize}
        \item model: The victim model used to generate the adversarial example.
        \item data: clean data.
        \item target: target label
        \item **kwargs: optional parameters.
    \end{itemize}
    
    \item \textbf{train}(self, train\_loader, optimizer, epoch)
    
    Call {\bf train} to train the adversarial model.
    \\[6pt]
    Parameters:
    \begin{itemize}
        \item train\_loader: the dataloader used to train robust model.
        \item optimizer: the optimizer for the training process.
        \item epoch: maximum epoch for the training process.
        \item **kwargs: optional input dependent on each derived class.
    \end{itemize}
    
    \item \textbf{test}(self, test\_loader)
    
    Call {\bf test()} function to test adversarial model.
    \\[6pt]
    Parameters:
    \begin{itemize}
        \item test\_loader: the dataloader used to test model.
    \end{itemize}
    
\end{itemize}

\subsubsection{Adversarial Training}
\textbf{deeprobust.image.defense.fgsmtraining} \ FGSM adversarial training ~\cite{goodfellow2014explaining} aims to improve model accuracy by training with adversarial examples. It generates adversarial examples in each iteration and updates model parameters via these adversarial examples.

~\\
\textbf{deeprobust.image.defense.fast} \ Fast ~\cite{wong2020fast} is an improved version of FGSM adversarial training. This work finds out that by simply adding a random initialization into the adversarial training samples' generating process, the model robustness would improve significantly

~\\
\textbf{deeprobust.image.defense.pgdtraining} \ PGD adversarial training uses adversarial examples generated by PGD instead of FGSM to train the model and achieve overall high performance.

~\\
\textbf{deeprobust.image.defense.YOPO} \ You-only-Propagate-Once(YOPO)~\cite{zhang2019you} is an accelerated version of the PGD adversarial training. When it generates the PGD adversarial examples for a $N$ layer network, it approximates the derivative of first $N-1$ layer as a constant, therefore there is no need to calculate the whole back propagation process in every iteration. Thus, the training time would be remarkably reduced.

~\\
\textbf{deprobust.image.defense.trades} \ The work~\cite{zhang2019theoretically} proposes a adversarial training strategy which encourages the clean samples and adversarial examples to be close in feature space. Its training objective is to minimize the loss:
\begin{equation}
     L(x_{trades}) = L(f(x),y) + \max_{||x'-x||\leq \epsilon} L(f(x),f(x'))
\end{equation}
This loss function can be devided into two parts, the first part is the natural loss while the second part set a goal for minimizing the distance between the classifier output for those examples that are close in input space.
Similar to the PGD adversarial training strategy, in each step, it first solves the inner maximization problem to find an optimal $x'$, and then updates model parameters to minimize the outside loss value. 

\subsubsection{Gradient Masking}
\textbf{deeprobust.image.defense.TherEncoding} \ Thermometer encoding~\cite{buckman2018thermometer} is one way to mask the gradient information of the DNN models, in order to avoid the attacker from finding successful adversarial examples. It uses a preprocessor to discretize an image's pixel value $x_i$ into a $l$-dimensional vector $\tau(x_i)$. (e.g. when $l = 10$, $\tau (0.66) = 1111110000$). The vector $\tau(x_i)$ acts as a ``thermometer'' to record the pixel $x_i$'s value.

\subsubsection{Detection}
\textbf{deeprobust.image.defense.LIDclassifier} Local Intrinsic Dimensionality(LID) detection ~\cite{ma2018characterizing} tries to train a classifier to distinguish adversarial examples from normal examples based on the LID features. Starting from a sample, it calculates the number of data points in a ball of a certain distance, and LID features measure the growth rate of the number of data points as the distance increases.

\section{Graph Package}
The design of graph package is slightly different from that of the image package. Specifically, graph package includes three main components, targeted attack, untargeted attack and defense. 
For these algorithms, supported networks and datasets are listed as follows:
\vskip 0.1in
Supported network:
\begin{itemize}
    \item GCN
\end{itemize}

Supported datasets:
\begin{itemize}
    \item Cora
    \item Cora-ml
    \item Citeseer
    \item Polblogs
    \item Pubmed
\end{itemize}

More details about adversarial attack and defense can be found in~\cite{jin2020adversarial}. In the following, we are going to illustrate the details of various subpackages.

\subsection{Targeted Attack Subpackage}
\textbf{deeprobust.graph.targeted\_attack}
\ This module introduces the API for targted attack methods in the graph package. In total, this package covers 5 algorithms: FGA~\cite{chen2018fga}, Nettack~\cite{nettack}, RL-S2V~\cite{rl-s2v}, IG-Attack~\cite{deep-insight-jaccard} and RND~\cite{nettack}. 

~\\
\textbf{deeprobust.graph.targeted\_attack.fga} \ FGSM~\cite{goodfellow2014explaining} can also be applied to attack graph data but it needs some modification to fit into the binary nature of graph data. One representative method to solve this problem is FGA~\cite{chen2018fga}. Basically, FGA first calculates the gradient of attack loss with respect to the graph structure and greedily chooses the perturbation with largest gradient.  

~\\
\textbf{deeprobust.graph.targeted\_attack.nettack} \ The work \cite{nettack} proposes an attack method called Nettack to generate structure and feature attacks on graphs. Nettack first selects possible perturbation candidates that would not violate degree distribution and feature co-occurrence of the original graph. Then it greedily chooses the perturbation that has the largest score to modify the graph. By doing this repeatedly until reaching the perturbation constraint, it can get the final modified graph. 

~\\
\textbf{deeprobust.graph.targeted\_attack.rl\_s2v} \ To do black-box query on the victim model, reinforcement learning is introduced. RL-S2V~\cite{rl-s2v} aims to employ the reinforcement learning technique to generate adversarial attacks on graph data under the black-box setting. It models the attack procedure as a Markov Decision Process (MDP) and the attacker is allowed to modify $m$ edges to change the predicted label of the target node $u$. Further, the Q-learning algorithm~\cite{mnih2013playing} is adopted to solve the MDP and guide the attacker to modify the graph. 

~\\
\textbf{deeprobust.graph.targeted\_attack.ig\_attack} \ 
Due to the discrete nature of graph data,
how to precisely approximate the gradient of adversarial perturbations is a big challenge. To solve this issue, IG attack~\cite{deep-insight-jaccard} suggests to use
integrated gradient~\cite{sundararajan2017axiomatic-ig} to better search for adversarial edges and feature perturbations. During
the attacking process, the attacker iteratively chooses the edge or feature which has the strongest
effect to the adversarial objective.  

~\\
\textbf{deeprobust.graph.targeted\_attack.rnd} \ RND is a baseline of attacking method used in~\cite{nettack}. Based on the assumption that unequal class labels are hindering classification, it modifies the graph structure sequentially. To be specific, given the target node, in each step it randomly samples nodes whose labels are different from the target node and then connects them in the graph. 

\subsection{Untargeted Attack Subpackage}

\textbf{deeprobust.graph.global\_attack}
\ This module introduces the API for untargted attack methods in the graph package. Currrently, this package covers 4 algorithms: Metattack~\cite{metattack}, PGD~\cite{xu2019topology-attack}, Min-max~\cite{xu2019topology-attack} and DICE~\cite{waniek2018hiding-dice}. 

~\\
\textbf{deeprobust.graph.global\_attack.metattack} 
\ Aiming to modify graph structure, Metattack~\cite{metattack} is a kind of untargeted poisoning attacks.  Basically, it treats the graph structure matrix as a hyper-parameter and calculates the meta gradient of the loss function with respect to graph structure. Further, A greedy approach is applied to select the perturbation based on the meta gradient.

~\\
\textbf{deeprobust.graph.global\_attack.topology\_attack} 
\ The work~\cite{xu2019topology-attack} considers two different settings: 1) attacking a fixed GNN and 2) attacking a re-trainable GNN. For attacking a fixed GNN, it utilizes the Projected Gradient Descent (PGD) algorithm in~\cite{madry2017towards} to search the optimal structure perturbation. This is called PGD attack. For the re-trainable GNNs, the attack problem is formulated as a min-max form where the inner maximization can be solved by gradient ascent and the outer minimization can be solved by PGD. It is called Min-max attack.

~\\
\textbf{deeprobust.graph.global\_attack.dice} \ DICE~\cite{waniek2018hiding-dice} means ``delete internally, connect externally" where it randomly connects nodes with different labels or drops edges between nodes sharing the same label. It is noted that DICE is a white-box attack and widely used as a baseline in comparing the performance of untargeted attacks.

\subsection{Defense Subpackage}

\subsubsection{Adversarial Training}
\textbf{deeprobust.graph.defense.adv\_training} 
\ Since adversarial training is a widely used countermeasure for adversarial attacks in the image data~\cite{goodfellow2014explaining}, we can also adopt this strategy to defend graph adversarial attacks.
The min-max optimization problem indicates that adversarial training involves two processes: (1) generating perturbations that maximize the prediction loss and (2) updating model parameters that minimize the prediction loss. By alternating the above two processes attractively, we can train a robust model against adversarial attacks. Since there are two inputs for graphs, i.e., adjacency matrix and attribute matrix, adversarial training can be done on them separately.

\subsubsection{Pre-processing}
\textbf{dedprobust.graph.defense.gcn\_jaccard}  \ The work~\cite{deep-insight-jaccard} proposes a preprocessing method based on two empirical observations of the attack methods: (1) Attackers usually prefer to adding edges over removing edges or modifying features and (2) Attackers tend to connect dissimilar nodes. Based on these findings, they propose a defense method by eliminating the edges whose two end nodes have small Jaccard Similarity~\cite{said2010social}.

\textbf{dedprobust.graph.defense.gcn\_svd} \ It is observed that Nettack~\cite{nettack} generates the perturbations which mainly change the small singular values of the graph adjacency matrix~\cite{entezari2020all-svd}. Thus it proposes to preprocess the perturbed adjacency matrix by using truncated SVD to get its low-rank approximation. 
 
\subsubsection{Attention Mechanism}
\textbf{deprobust.graph.defense.rgcn} 
\ Different from the above preprocessing methods which try to exclude adversarial perturbations, RGCN~\cite{rgcn} aims to train a robust GNN model by penalizing model's weights on adversarial edges or nodes. Based on the assumption that adversarial nodes may have high prediction uncertainty, they propose to model the hidden representation of nodes as Gaussian distribution with mean value and variance where the uncertainty can be reflected in the variance. When aggregating the information from neighbor nodes, it applies an attention mechanism to penalize the nodes with high variance.

\section{Hand-on Case Studies}
In this section, we would give concrete examples to illustrate how to use this repository. For each type of methods, we provide one demo code.

\subsection{Image Case Studies}
\subsubsection{Train Network}
In {\bf deeprobust.image.netmodels}, we provide several deep network architecture. Call {\bf train()} to train a model.
\begin{lstlisting}
import deeprobust.image.netmodels.train_model as trainmodel
trainmodel.train('CNN', 'MNIST', 'cuda', 20)
\end{lstlisting}

\subsubsection{Attack}
To launch an attack method, The first step is to import certain attack class from {\bf deeprobust.image.attack}. Then, we need to initialize a victim model and create a dataloader, which contains the test images to be generated as adversarial examples. Then, we can feed the model and data to the attack method. The output would be adversarial examples. 
\begin{lstlisting}
from deeprobust.image.attack.pgd import PGD
from deeprobust.image.config import attack_params
import deeprobust.image.netmodels.resnet as resnet

model = resnet.ResNet18().to('cuda')
model.load_state_dict(torch.load \
                     ("./trained_models/MNIST_CNN_epoch_20.pt"))
model.eval()

transform_val = transforms.Compose([transforms.ToTensor()])
test_loader  = torch.utils.data.DataLoader(
                datasets.CIFAR10('deeprobust/image/data', 
                train=False, download=True, transform=transform_val),
                batch_size=10, shuffle=True)

x, y = next(iter(test_loader))
x = x.to('cuda').float()

adversary = PGD(model, device)
Adv_img = adversary.generate(x, y, **attack_params['PGD_CIFAR10'])
\end{lstlisting}

\subsubsection{Defense}
Defense method can be imported in {\bf deeprobust.image.defense}. We need to feed a model structure and a dataloader to the defense model. The output would be adversarial trained model and the performance on both clean data and adversarial data. 

\begin{lstlisting}
from deeprobust.image.defense.pgdtraining import PGDtraining
from deeprobust.image.config import defense_params
from deeprobust.image.netmodels.CNN import Net
import torch
from torchvision import datasets, transforms 

model = Net()
train_loader = torch.utils.data.DataLoader(
               datasets.MNIST('deeprobust/image/defense/data', 
               train=True, download=True,
               transform=transforms.Compose([transforms.ToTensor()])),
               batch_size=100,shuffle=True)

test_loader = torch.utils.data.DataLoader(
              datasets.MNIST('deeprobust/image/defense/data', 
              train=False,
              transform=transforms.Compose([transforms.ToTensor()])),
                batch_size=1000,shuffle=True)

defense = PGDtraining(model, 'cuda')
defense.generate(train_loader, test_loader, \    
                 **defense_params["PGDtraining_MNIST"])
\end{lstlisting}

\subsubsection{Evaluation}
We provide a simple access to evaluate the performance of attack toward defense. 
\begin{lstlisting}
cd DeepRobust
#creat a victim model
python examples/image/test_train.py
#evaluation attack 
python deeprobust/image/evaluation_attack.py --attack_method PGD 
--attack_model CNN --dataset MNIST
\end{lstlisting}

\subsection{Graph Case Studies}
\subsubsection{Attack Graph Neural Networks}
We show an example of attacking graph neural networks. We will use a linearized GCN as the surrogate model and apply untargeted Metattack to generate perturbed graph on the Cora citation dataset.

First we need to import the packages we are going to use in the head of the code and load Cora dataset. 
\begin{lstlisting}
import torch
from deeprobust.graph.data import Dataset
from deeprobust.graph.defense import GCN
from deeprobust.graph.global_attack import Metattack

# load dataset
data = Dataset(root='/tmp/', name='cora', setting='nettack')
adj, features, labels = data.adj, data.features, data.labels
idx_train,idx_val,idx_test = data.idx_train,data.idx_val,data.idx_test
\end{lstlisting}

Then set up the surrogate model to be attacked.
\begin{lstlisting}
# set up surrogate model
device = torch.device("cuda:0" \
                      if torch.cuda.is_available() else "cpu")
surrogate = GCN(nfeat=features.shape[1], nclass=labels.max().item()+1, 
                nhid=16,with_relu=False, device=device)
surrogate = surrogate.to(device)
surrogate.fit(features, adj, labels, idx_train)
\end{lstlisting}

Then we use Metattack to generate perturbations to attack the surrogate model. Here the variable \textit{modified\_adj} is the perturbed graph generated by Metattack. 

\begin{lstlisting}
# use Metattack to generate attacks
model = Metattack(surrogate, nnodes=adj.shape[0],
                  feature_shape=features.shape, device=device)
model = model.to(device)
perturbations = int(0.05 * (adj.sum() // 2))  # set attack budget
model.attack(features, adj, labels, idx_train, 
             idx_unlabeled, perturbations)
modified_adj = model.modified_adj
\end{lstlisting}

\subsubsection{Defend Graph Adversarial Attacks}
We show an example of defending graph adversarial attacks. We will use Metattack as the attacking method and GCN-Jaccard as the defense method. 

First, we import all the packages we need to use and load the clean graph and pre-attacked graph of Cora dataset.

\begin{lstlisting}
import torch
from deeprobust.graph.data import Dataset, PtbDataset
from deeprobust.graph.defense import GCN, GCNJaccard
import numpy as np

np.random.seed(15)
# load clean graph
data = Dataset(root='/tmp/', name='cora', setting='nettack')
adj, features, labels = data.adj, data.features, data.labels
idx_train,idx_val,idx_test = data.idx_train,data.idx_val,data.idx_test

# load pre-attacked graph by mettack
perturbed_data = PtbDataset(root='/tmp/', name='cora')
perturbed_adj = perturbed_data.adj
\end{lstlisting}

Then we set up the defense model GCN-Jaccard and test it performance on the perturbed graph.

\begin{lstlisting}
# Set up defense model and test performance 
device = torch.device("cuda:0" \
                      if torch.cuda.is_available() else "cpu")
model = GCNJaccard(nfeat=features.shape[1], nclass=labels.max()+1, 
                   nhid=16, device=device)         
model = model.to(device)
model.fit(features, perturbed_adj, labels, idx_train)
model.eval()
output = model.test(idx_test)
\end{lstlisting}

As a comparison, we can also set up GCN model and test its performance on the perturbed graph.

\begin{lstlisting}
# Test GCN on the perturbed graph
model = GCN(nfeat=features.shape[1], nclass=labels.max()+1, 
            nhid=16, device=device)
model = model.to(device)
model.fit(features, perturbed_adj, labels, idx_train)
model.eval()
output = model.test(idx_test)
\end{lstlisting}

\section{Conclusion}
Our main goal is to provide a comprehensive, easy-to-use platform for researchers who are interested in adversarial attack and defense. In the future, we would support larger datasets and more model architectures. Moreover, we will keep including the newest models and updating this repository.
\newpage
\bibliographystyle{unsrt}
\bibliography{References}

\newpage
\begin{appendices}
\section{Environment Dependencies}

\begin{table}[h]
\centering
\Large
\begin{tabular*}{0.6\textwidth}{p{0.4\textwidth}|p{0.2\textwidth}}
\toprule
Dependency    & Version \\ \midrule
torch         & $\geq$1.2.0   \\
torchvision   & $\geq$0.4.0   \\
numpy         & $\geq$1.17.1  \\
matplotlib    & $\geq$3.1.1   \\
scipy         & $\geq$1.3.1   \\
Pillow        & $\geq$7.0.0   \\
scikit\_learn & $\geq$0.22.1  \\
skimage       &  0     \\
tensorboardX  & $\geq$2       \\
tqdm          & $\geq$4.42.1  \\
texttable     & $\geq$1.6.2   \\
numba         & $\geq$0.48.0 \\
\bottomrule
\end{tabular*}
\caption{Dependencies}

\end{table}

\section{Structure tree}
\begin{lstlisting}[language={[ANSI]C}, basicstyle=\ttfamily]
|___    LICENSE
|___    README.md
|___    adversary_examples
|___    deeprobust
|   |___    __init__.py
|   |___    graph
|   |   |___    README.md
|   |   |___    __init__.py
|   |   |___    black_box.py
|   |   |___    data
|   |   |   |___    __init__.py
|   |   |   |___    attacked_data.py
|   |   |   |__ dataset.py
|   |   |___    defense
|   |   |   |___    __init__.py
|   |   |   |___    adv_training.py
|   |   |   |___    gcn.py
|   |   |   |___    gcn_preprocess.py
|   |   |   |___    r_gcn.py
|   |   |   |__ r_gcn.py.backup
|   |   |___    examples
|   |   |   |___    test_adv_train_evasion.py
|   |   |   |___    test_adv_train_poisoning.py
|   |   |   |___    test_dice.py
|   |   |   |___    test_fgsm.py
|   |   |   |___    test_gcn.py
|   |   |   |___    test_gcn_jaccard.py
|   |   |   |___    test_gcn_svd.py
|   |   |   |___    test_mettack.py
|   |   |   |___    test_nettack.py
|   |   |   |___    test_nipa.py
|   |   |   |___    test_random.py
|   |   |   |___    test_rgcn.py
|   |   |   |___    test_rl_s2v.py
|   |   |   |___    test_rnd.py
|   |   |   |__ test_topology_attack.py
|   |   |___    global_attack
|   |   |   |___    __init__.py
|   |   |   |___    base_attack.py
|   |   |   |___    dice.py
|   |   |   |___    mettack.py
|   |   |   |___    nipa.py
|   |   |   |___    random.py
|   |   |   |__ topology_attack.py
|   |   |___    requirements.txt
|   |   |___    rl
|   |   |   |___    env.py
|   |   |   |___    nipa.py
|   |   |   |___    nipa_config.py
|   |   |   |___    nipa_env.py
|   |   |   |___    nipa_nstep_replay_mem.py
|   |   |   |___    nipa_q_net_node.py
|   |   |   |___    nstep_replay_mem.py
|   |   |   |___    q_net_node.py
|   |   |   |___    rl_s2v.py
|   |   |   |___    rl_s2v_config.py
|   |   |   |__ rl_s2v_env.py
|   |   |___    targeted_attack
|   |   |   |___    __init__.py
|   |   |   |___    base_attack.py
|   |   |   |___    evaluation.py
|   |   |   |___    fgsm.py
|   |   |   |___    nettack.py
|   |   |   |___    rl_s2v.py
|   |   |   |__ rnd.py
|   |   |__ utils.py
|   |__image
|       |___    README.md
|       |___    __init__.py
|       |___    adversary_examples
|       |___    attack
|       |   |___    BPDA.py
|       |   |___    Nattack.py
|       |   |___    Universal.py
|       |   |___    YOPOpgd.py
|       |   |___    __init__.py
|       |   |___    base_attack.py
|       |   |___    cw.py
|       |   |___    deepfool.py
|       |   |___    fgsm.py
|       |   |___    l2_attack.py
|       |   |___    lbfgs.py
|       |   |___    onepixel.py
|       |   |__ pgd.py
|       |___    config.py
|       |___    data
|       |___    defense
|       |   |___    LIDclassifier.py
|       |   |___    TherEncoding.py
|       |   |___    YOPO.py
|       |   |___    __init__.py
|       |   |___    advexample_pgd.png
|       |   |___    base_defense.py
|       |   |___    fast.py
|       |   |___    fgsmtraining.py
|       |   |___    pgdtraining.py
|       |   |___    test_PGD_defense.py
|       |   |___    trade.py
|       |   |__ trades.py
|       |___    evaluation_attack.py
|       |___    netmodels
|       |   |___    CNN.py
|       |   |___    CNN_multilayer.py
|       |   |___    YOPOCNN.py
|       |   |___    __init__.py
|       |   |___    resnet.py
|       |   |___    train_model.py
|       |   |__ train_resnet.py
|       |___    optimizer.py
|       |___    synset_words.txt
|       |__ utils.py
|___    examples
|   |___    graph
|   |   |___    test_adv_train_evasion.py
|   |   |___    test_adv_train_poisoning.py
|   |   |___    test_dice.py
|   |   |___    test_fgsm.py
|   |   |___    test_gcn.py
|   |   |___    test_gcn_jaccard.py
|   |   |___    test_gcn_svd.py
|   |   |___    test_mettack.py
|   |   |___    test_nettack.py
|   |   |___    test_nipa.py
|   |   |___    test_random.py
|   |   |___    test_rgcn.py
|   |   |___    test_rl_s2v.py
|   |   |__ test_rnd.py
|   |__ image
|       |___    __init__.py
|       |___    __pycache__
|       |   |___    __init__.cpython-36.pyc
|       |   |__ test_cw.cpython-36.pyc
|       |___    test1.py
|       |___    test_PGD.py
|       |___    test_cw.py
|       |___    test_deepfool.py
|       |___    test_fgsm.py
|       |___    test_lbfgs.py
|       |___    test_nattack.py
|       |___    test_onepixel.py
|       |___    test_pgdtraining.py
|       |___    test_trade.py
|       |___    test_train.py
|       |__ testprint_mnist.py
|___    get-pip.py
|___    requirements.txt
|___    setup.py
|___    tree.md
|__ tutorials
    |___    __init__.py
    |___    test1.py
    |___    test_PGD.py
    |___    test_cw.py
    |___    test_deepfool.py
    |___    test_fgsm.py
    |___    test_lbfgs.py
    |___    test_nattack.py
    |___    test_onepixel.py
    |___    test_pgdtraining.py
    |___    test_trade.py
    |___    test_train.py
    |__ testprint_mnist.py
\end{lstlisting}

\end{appendices}

\end{document}